\documentclass{l4dc2025}
\usepackage[T1]{fontenc}

\title[Anti-fragility in RL]{\textcolor{black}{
Parameter Stress Analysis in Reinforcement Learning: Applying Synaptic Filtering to Policy Networks}
}
\usepackage{times}

\author{%
 \Name{Zain ul Abdeen} \thanks{Correspondence to: Zain ul Abdeen {<zabdeen@vt.edu>}} \\
 \addr Bradley Department of Electrical and Computer Engineering\\
       Virginia Tech
 \AND
 \Name{Ming Jin}\\
 \addr Bradley Department of Electrical and Computer Engineering\\
       Virginia Tech
}
\makeatletter
\def\@jmlrvolume{}   
\def\jmlrproceedings#1#2{} 
\makeatother

\begin{document}

\maketitle

\begin{abstract}%

This paper explores reinforcement learning (RL) policy robustness by systematically analyzing network parameters under internal and external stresses. \textcolor{black}{We apply synaptic filtering methods using high-pass, low-pass, and pulse-wave filters from} \citep{pravin2024fragility}, as an internal stress by selectively perturbing parameters, while adversarial attacks apply external stress through modified agent observations. This dual approach enables the classification of parameters as \textit{fragile}, \textit{robust}, or \textit{antifragile}, based on their influence on policy performance in clean and adversarial settings. Parameter scores are defined to quantify these characteristics, and the framework is validated on proximal policy optimization (PPO)-trained agents in Mujoco continuous control environments. The results highlight the presence of antifragile parameters that enhance policy performance under stress, demonstrating the potential of targeted filtering techniques to improve RL policy adaptability. These insights provide a foundation for future advancements in the design of robust and antifragile RL systems.
\end{abstract}


\section{Introduction}
Reinforcement learning has demonstrated broad success across diverse domains \citep{gu2024review}. However, RL agents exhibit vulnerabilities to various perturbations, highlighting the need for a deeper understanding of their capacity to adapt and generalize in dynamic and adversarial environments. 

\begin{figure}
\centering
\includegraphics[width=0.8\linewidth]{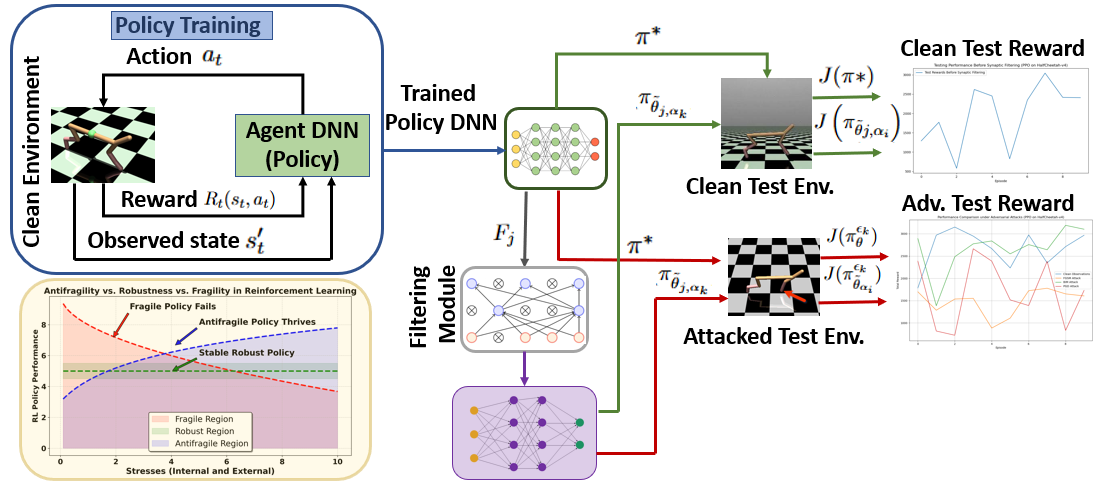}
\caption{Framework for training and evaluating robust and antifragile policies in RL.}
\label{baseload}
\end{figure}

\textcolor{black}{This study applies the synaptic filtering framework for neural network analysis developed by} \citep{pravin2024fragility} \textcolor{black}{to RL policies. We adopt their methodology including: (i) \textit{internal stress}, involving parameter perturbations through their three specific filter types (high-pass, low-pass, and pulse-wave filters) to RL policy network, (ii) \textit{external stress}, induced by adversarial perturbations to the agent observations}, \textcolor{black}{and (iii) their parameter scoring scheme to classify parameters as \textit{fragile, robust,} or \textit{antifragile}}. Parameters critical to performance degradation are labeled as \textit{fragile}, while those unaffected by stress are identified as \textit{robust}. Parameters that contribute to improved performance under stress are classified as \textit{antifragile}, providing insights into the resilience and adaptability of RL agents. Figure~\ref{baseload} illustrates this categorization, showing how RL agent performance varies under different levels of stress. Our contribution is demonstrating this framework, originally developed for supervised learning, can be applied to RL policies where we use cumulative rewards instead of classification accuracy as the performance metric.


We validate the approach using agents trained with the PPO algorithm \citep{schulman2017proximal} in continuous control environments from the OpenAI Gym benchmark suite \citep{brockman2016openai}, which provide diverse challenges for evaluating RL policy robustness. Our results show that introducing controlled stressors reveals parameter vulnerabilities and identifies antifragile characteristics, offering a pathway for designing more resilient and adaptive RL systems. \textcolor{black}{The key contributions of this work is: We show that} \citep{pravin2024fragility} \textcolor{black}{synaptic filtering framework can be directly applied to RL policies trained with PPO, revealing similar parameter fragility patterns when using cumulative rewards as the performance metric.}

\section{Related work}
 Recently, antifragility has gained attention in robotics~\citep{axenie2023antifragile}. \textcolor{black}{In the context of deep learning,} \citep{pravin2024fragility} \textcolor{black}{introduced a systematic framework for characterizing neural network parameters as fragile, robustness, and antifragile through synaptic filtering. Their methodology uses three specific filter types (high-pass, low-pass, and pulse-wave) to systematically perturb network parameters as internal stress, combined with adversarial attacks as external stress. They define parameter scores based on performance deviation from baseline to classify parameters. We directly adopt their framework and apply it to RL policies, using cumulative rewards instead of classification accuracy.} 

 Related work on network pruning has shown that removing certain parameters can sometimes improve performance \citep{molchanov2019importance}, suggesting that not all parameters contribute equally to network function. Several pruning-based approaches~\citep{ramanujan2020s} and diversity-driven filtering strategies~\citep{mariet2015diversity} have demonstrated that structural modification of neural networks can lead to performance gains. 
 \textcolor{black}{In the RL domain, adversarial robustness has been studied primarily through gradient-based attacks} \citep{huang2017adversarial}, \textcolor{black}{revealing policy vulnerabilities to observation perturbations. However, the systematic application of the parameter characterization framework proposed by} \citep{pravin2024fragility} \textcolor{black}{has not previously been explored in RL. Our work bridges this gap by demonstrating that their methodology reveals similar fragility patterns in sequential decision-making contexts.}

\section{Preliminaries}

\subsection{Markov decision process (MDP)} RL is a computational framework where an agent learns to make decisions by interacting with an environment to maximize cumulative rewards. Formally, RL is modeled as a finite MDP, represented by the tuple $\mathcal{M}=\left(\mathcal{S},\mathcal{A},T,P,R\right)$, where $\mathcal{S}$ and $\mathcal{A}$ are the state and action spaces, respectively, and $T$ is the horizon length. The environment dynamics are defined by transition function $P=\left\lbrace P_{t}(s'_{t}|s_{t},a_{t})\right\rbrace_{t=1}^{T}$, which gives the probability of transition from state $s_{t}\in \mathcal{S}$ to $s'_{t}\in \mathcal{S}$ given action $a_{t}\in \mathcal{A}$ at time step $t$, and $R=\left\lbrace R_{t}(s_{t},a_{t})\right\rbrace_{t=1}^{T}$ denotes the immediate reward function.  The agent follows a policy $\pi:\mathcal{S}\times \mathcal{A}\to [0,1]$, where $\pi\left(a|s\right)$ is the probability of selecting action $a$ in state $s$. 
The goal is to find an optimal policy $\pi^{*}$ that maximizes the expected cumulative  reward $J(\pi)$ from any initial state $s_{0}$:
\begin{equation}\label{eq1}
    \pi^{*}_{a_{t}|s_{t}}=\arg\max_{\pi}J(\pi)=\arg\max_{\pi}\mathbb{E}_{a_{t}\sim\pi}\left[\sum_{t=0}^{T}R_{t}(s_{t},a_{t})\right]
\end{equation}
We employ policy-based RL algorithm, which directly optimize in the policy space to achieve the maximization in \eqref{eq1}, and policy is instantiated using a deep neural network, i.e., $\pi_{\theta}\left(a_{t}|s_{t}\right)$, where~$\theta$ represents the parameters (weights and biases) of the policy network. Following a given policy $\pi_{\theta}\left(a_{t}|s_{t}\right)$, the RL controller performance can be written as $J(\theta)=\mathbb{E}_{a_{t}\sim\pi_{\theta}}\left[\sum_{t=0}^{T}R_{t}(s_{t},a_{t})\right]$. Therefore, objective becomes optimizing policy parameters, i.e., $\theta^{*}=\arg\max_{\theta}J(\theta)$. To this end, we use the PPO algorithm, use gradient ascent for policy update: $\theta_{t+1}=\theta_{t}+\eta\hat{\nabla}_{\theta}J(\theta)$, 
where $\hat{\nabla}_{\theta}J(\theta)$ is the policy gradient estimated from collected experience, and $\eta$ is the learning rate.

\subsection{External Stress }
External stress in RL refers to factors that negatively impact an agent performance by disrupting its interaction with the environment, such as changes in observations, actions, or environment dynamics.  
In our investigation, Fast Gradient Sign Method (FGSM) \citep{goodfellow2015explaining} is utilized, a computationally efficient technique for generating adversarial examples through single-step perturbations. FGSM creates these perturbations by modifying the input state based on the gradient of the loss function. For an observation state~$s_t$, an adversarial perturbation~$\delta_{\epsilon}$ of magnitude $\epsilon$ is calculated as: $\delta_{\epsilon}=\epsilon\cdot \text{sign} \left(\nabla_{s_{t}}J(\theta,s_{t})\right)$. 
The perturbation observation is then $s_{t}^{\epsilon}=s_{t}+\delta_{\epsilon}$. 
The loss function used for generating the adversarial perturbation is defined as $J(\theta,s_t)=-\log\pi_{\theta}(a_{t}|s_{t})$. Maximizing this loss reduces the agent confidence in its chosen action, potentially leading to suboptimal or incorrect decisions.
External stress is applied with various magnitudes $\epsilon={ \epsilon_0, \epsilon_1, \dots, \epsilon_M }$, ranging from a minimum $\epsilon_0$ to a maximum perturbation magnitude $\epsilon_M$ with a step size $\Delta \epsilon$. This results in a set of perturbed observations $\mathcal{S}_\epsilon= [ s_t^{\epsilon_0}, s_t^{\epsilon_1}, \dots, s_t^{\epsilon_M} ]$ for the agent. The set of policy networks under external stress defined as:
\begin{equation} \Pi^\epsilon = [ \pi_\theta( a_t | s_t^{\epsilon_0} ), \pi_\theta( a_t | s_t^{\epsilon_1} ), \dots, \pi_\theta( a_t | s_t^{\epsilon_M} ) ]. \end{equation}
Analyzing the performance of the agent under varying levels of external stress assesses policy robustness to external stress. The cumulative rewards obtained under different magnitudes as:
\begin{equation}
    \mathcal{J}^{\epsilon}=[J(\pi_{\theta}^{\epsilon_{0}}),J(\pi_{\theta}^{\epsilon_{1}}),\dots,J(\pi_{\theta}^{\epsilon_{M}})]
\end{equation}

Examining how $\mathcal{J}^\epsilon$ varies with increasing $\epsilon$ quantifies performance degradation due to adversarial attacks. This approach identifies the specific perturbation magnitudes required to significantly degrade performance or cause agent failure.

\section{Methodology for Policy Characterization}

In this section, we detail the methodology employed to characterize the policy parameters as fragile, robust, or antifragile under internal and external stress. \textcolor{black}{Our approach builds on the synaptic filtering framework of} \citep{pravin2024fragility} \textcolor{black}{as internal stress, which was originally proposed for supervised deep learning, and we extends it to the policy network of an RL agent.} This section first explain how we apply the internal stress and then introduce parameter scores that characterize the parameters. 


\subsection{Internal Stress}
Internal stress involves perturbing the parameters of policy network $\pi_{\theta}(\cdot)$ to evaluate their impact on performance. We implement this through \textbf{synaptic filtering}, which systematically modifies or removes parameters based on their magnitudes.
Formally, synaptic filtering in RL is defined as: ``Given a policy network $\pi_{\theta}(\cdot)$ with parameters $\theta \in \mathbf{R}^{n}$, synaptic filtering applies a mask $m_{\alpha} \in \left\lbrace0,1\right\rbrace^{n}$ determined by a filtering function $F_{\alpha}(\cdot)$. \textcolor{black}{The filtered parameters $\tilde{\theta}$ identical to those in} \citep{pravin2024fragility}:
\begin{equation} \tilde{\theta} = F_{\alpha}(\pi_{\theta})= m_{\alpha} \odot \theta, \label{eq5} \end{equation} 
where $\odot$ denotes the element-wise Hadamard product''. 
The mask $m_{\alpha}$ is determined by the filtering method and threshold $\alpha =\left\lbrace\alpha_0, \alpha_1, \dots, \alpha_N\right\rbrace$ represent a normalized set of synaptic filtering thresholds spanning the entire parameter range with lower bound $\alpha_0 = \min \left\lbrace
|\theta|\right\rbrace$, upper bound $\alpha_N = \max \left\lbrace
|\theta|\right\rbrace$, and step size $\Delta \alpha=\frac{\alpha_{N}-\alpha_{0}}{N}$, such that  $\alpha_i = \alpha_{i-1} + \Delta \alpha$. For meaningful analysis, all policy parameters satisfies ($\theta \neq 0$); otherwise, the policy network $\pi_\theta(\cdot)$ produces invalid outputs.

\subsubsection{Synaptic Filtering Methods}
We employ the three synaptic filtering methods introduced by \citep{pravin2024fragility}. The mathematical formulation for the hight-pass filter ($F_{HPF}$), low-pass filter ($F_{LPF}$), and pulse-wave filter $F_{PWF}$ are adopted directly from their work \citep[Eqs. 7-9]{pravin2024fragility} as follows:\\
\textbf{High-Pass Filter $(F_{HPF})$ :} This filter removes parameters with absolute values below a threshold $\alpha_i$. 
\begin{equation}
\tilde{\theta}_{1,\alpha_i} = F_{HPF}(\theta, \alpha_i) =
\begin{cases} 
0, & \text{if } |\theta| \leq \alpha_i,  \\
\theta, & \text{otherwise}.
\end{cases}
\end{equation}
\\
\textbf{Low-Pass Filter ($F_{LPF}$)}:
The filter removes parameters with absolute values above a threshold $\alpha_i$. \begin{equation}
\tilde{\theta}_{2,\alpha_i} = F_{LPF}(\theta, \alpha_i) =
\begin{cases} 
0, & \text{if } |\theta| \geq \alpha_i,  \\
\theta, & \text{otherwise}.
\end{cases}
\end{equation} 
\\
\textbf{Pulse-Wave Filter ($F_{PWF}$)}: This filter removes parameters within a narrow band around the threshold $\alpha_i$.
\begin{equation} \tilde{\theta}_{3,\alpha_i} = F_{PWF}(\theta, \alpha_i) = \begin{cases} 0, & \text{if } \alpha_i - \frac{\Delta \alpha}{2} < |\theta| \leq \alpha_i + \frac{\Delta \alpha}{2}, \\ \theta, & \text{otherwise}. 
\end{cases} \end{equation} 

\begin{figure} 
    \begin{minipage}{0.32\textwidth}
        \includegraphics[width=\textwidth]{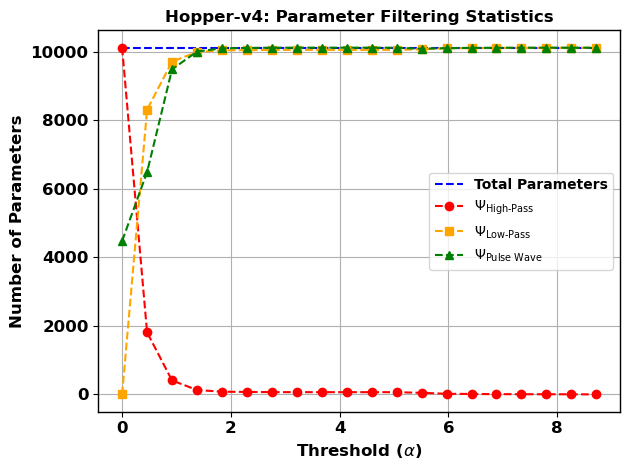}
    \end{minipage}
    \hfill
    \begin{minipage}{0.32\textwidth}
        \includegraphics[width=\textwidth]{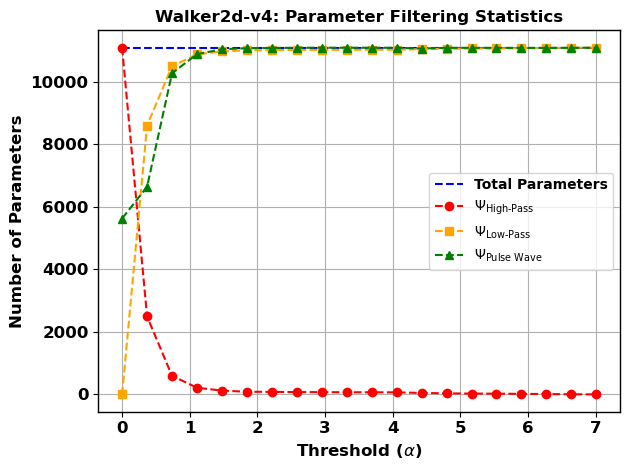}
    \end{minipage}
    \hfill
    \begin{minipage}{0.32\textwidth}
        \includegraphics[width=\textwidth]{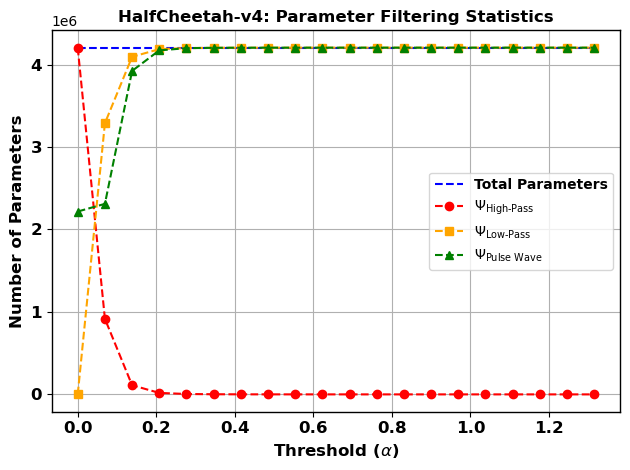}
    \end{minipage}
    \caption{Global parameter filtering statics for mujoco environment}
    \label{parameterstat}
\end{figure}

For each filter $F_{j}$, we define the  \textit{network compactness}: $\Psi_{j, \alpha_{i}}= 1-\frac{\psi_{j,\alpha_{i}}}{\psi}$ as the portion of parameter retained after filtering, where $\psi$ is the total number of parameters and $\psi_{j,\alpha_{i}}$ 
  is the number of parameters filtered out by filter.
The Figure \eqref{parameterstat} illustrates the distribution of parameter magnitudes, showing that the high-pass filter removes most parameters at low thresholds, indicating a concentration of small-magnitude parameters.
By systematically varying $\alpha_{i}$, these filters generate three sets of perturbed parameters: 
\begin{equation} \tilde{\Theta}_1 = \{ \tilde{\theta}_{1,\alpha_i} ~~\text{for}~~ \alpha_i \in \alpha \}, \quad \tilde{\Theta}_2 = \{ \tilde{\theta}_{2,\alpha_i} ~~\text{for}~~ \alpha_i \in \alpha \}, \quad \tilde{\Theta}_3 = \{ \tilde{\theta}_{3,\alpha_i} ~~\text{for}~~ \alpha_i \in \alpha \}. \end{equation}
Correspondingly, we obtain three sets of perturbed policy networks: \begin{equation} \Pi_1 = \{ \pi_{\tilde{\theta}_{1,\alpha_i}}(\cdot) \mid \alpha_i \in \alpha \}, \quad \Pi_2 = \{ \pi_{\tilde{\theta}_{2,\alpha_i}}(\cdot) \mid \alpha_i \in \alpha \}, \quad \Pi_3 = \{ \pi_{\tilde{\theta}_{3,\alpha_i}}(\cdot) \mid \alpha_i \in \alpha \}. \end{equation}

By applying these filters to the policy parameters, we assess 
the agent ability to make decisions and achieve rewards when its internal parameters are perturbed. This analysis helps us understand the sensitivity of the policy performance to changes in different regions of the parameter space. For each filter $F_j$ (where $j \in \{\text{HPF}, \text{LPF}, \text{PWF}\}$),
we measure the cumulative rewards: 

\begin{equation} \mathcal{J}_{j,\alpha}= [J(\pi_{\tilde{\theta}_{j,\alpha_{0}}}),J(\pi_{\tilde{\theta}_{j,\alpha_{1}}}),\dots,J(\pi_{\tilde{\theta}_{j,\alpha_{k}}})], \end{equation}

where, $J\left(\pi_{\tilde{\theta}{j,\alpha_i}}\right)$ represents the cumulative reward achieved by perturbed policy $\pi_{\tilde{\theta}_{j,\alpha_i}}$. Comparing these results to unperturbed policy performance $J(\pi_{\theta})$ quantifies the impact of internal stress. 
\subsection{Parameter Characterization}
In this subsection we define the three characterizations of policy parameters as: \textbf{\textit{fragility}}, \textbf{\textit{robustness}} and \textbf{\textit{antifragility}}. \textcolor{black}{In order to define these characterization, following} \citep{pravin2024fragility} \textcolor{black}{parameter scoring concept, we define parameter score as metric which measure the performance of perturbed and unperturbed policy networks.}


\subsubsection{Parameter Score}

The \textit{parameter score} quantifies the effect of stress on policy performance. 
It is defined as the difference in performance between the stressed policy and the baseline policy reward,
\begin{equation} S \approx \int_{\sigma_{min}}^{\sigma_{max}} J_{\text{stressed}}(\sigma) - J_{\text{baseline}}(\sigma)d \sigma,  \end{equation}

where, $J_{\text{stressed}}$ is the expected cumulative reward under stress, either $J(\pi_{\tilde{\theta}{\alpha_i}})$ for internal stress or $J(\pi_{\theta}^{\epsilon_j})$ for external stress.  $J_{\text{baseline}} = J(\pi_{\theta})$ is the expected cumulative reward of the baseline, unperturbed policy. 
The parameter score $S$ thus captures the net effect of stress on cumulative reward received by the agent. 
\paragraph{Parameter Score for Clean Environment:} This metric measures the impact of synaptic filtering under clean environment at different thresholds $\alpha_{i}$ on the agent performance. It is computed as:

\begin{equation} S_{\alpha_i} = J(\pi_{\tilde{\theta}_{\alpha_i}}) - J(\pi_{\theta}).  \end{equation}

 Assessing the parameter score in a clean environment allows us to assess the intrinsic sensitivity of the policy network to internal perturbations. By systematically filtering parameters and observing the resultant change in performance, we can identify which parameters are critical for maintaining performance \textit{(robust)}, which negatively impact performance when perturbed \textit{(fragile)}, and which enhance performance upon modification \textit{(antifragile)}. This metric provides a foundational understanding of parameter importance and resilience within the network's architecture.
\\
\textbf{Parameter Score Under Adversarial Environment:}
When external stress is introduced via adversarial attacks with perturbation magnitude $\epsilon_k$, the parameter score is calculated as:

\begin{equation} S^{\epsilon_k} = J(\pi_{\tilde{\theta}_{\alpha_i}}^{\epsilon_{k}}) - J(\pi^{\epsilon_{k}}_{\theta}).  \end{equation}

Evaluating the parameter score under adversarial environments allows us to understand how synaptic filtering influences the resilience of policy network to external perturbations. This analysis reveals whether certain parameters not only maintain performance under normal conditions but also enhance or diminish robustness against adversarial attacks.\\
\textbf{Combined Difference of Parameter Scores:} To evaluate the interplay between internal and external stresses, we compute the combined parameter score:

\begin{equation} \Delta S^{\epsilon_k}_{\alpha_{i}} = J(\pi_{\tilde{\theta}_{\alpha_i}}^{\epsilon_k}) - J(\pi_{\tilde{\theta}_{\alpha_i}}). \label{eq
} \end{equation}
The combined parameter score captures how adversarial attacks affect a synaptically filtered network compared to its clean, filtered counterpart. This metric highlights the differential impact of adversarial stress on the network after internal perturbations, providing a nuanced understanding of parameter resilience and adaptability. It helps identify parameters that not only sustain performance under internal stress but also enhance or diminish robustness against external threats.

\section{Experimental Framework and Antifragility Analysis of RL Policies}
\subsection{Experiment Setup}

We conducted experiments on three continuous control environments from the Gymnasium library: \texttt{Walker2D}-v4, \texttt{Hopper}-v4, and \texttt{HalfCheetah}-v4. These environments were selected for their high-dimensional state spaces and challenging control tasks, making them suitable for testing policy robustness and antifragility. The RL policies were trained using the PPO algorithm, implemented with the Stable-Baselines3 framework \citep{raffin2021stable}. Both the policy and value networks used a multilayer perceptron (MLP) architecture with three hidden layers containing 512, 256, and 128 neurons, activated by ReLU function. Training process was conducted with a learning rate of $1\times 10^{-4}$
  and a batch size of 128. Performance was evaluated using cumulative rewards over multiple episodes. Once trained, the policies were subjected to two types of stress: (i) \textit{internal stress}, induced via synaptic filtering and (ii) \textit{external stress}, introduced through adversarial perturbations. The goal was to analyze the fragility, robustness, and antifragility of policy parameters under these conditions.
\begin{figure}
    \begin{minipage}{0.32\textwidth}
        \includegraphics[width=\textwidth]{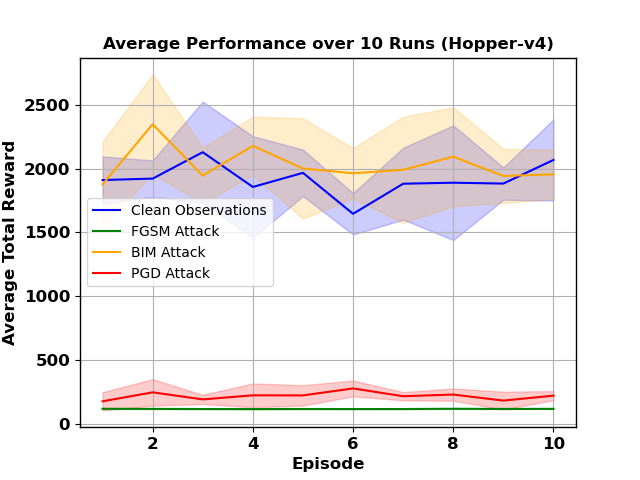}
    \end{minipage}
    \hfill
    \begin{minipage}{0.32\textwidth}
        \includegraphics[width=\textwidth]{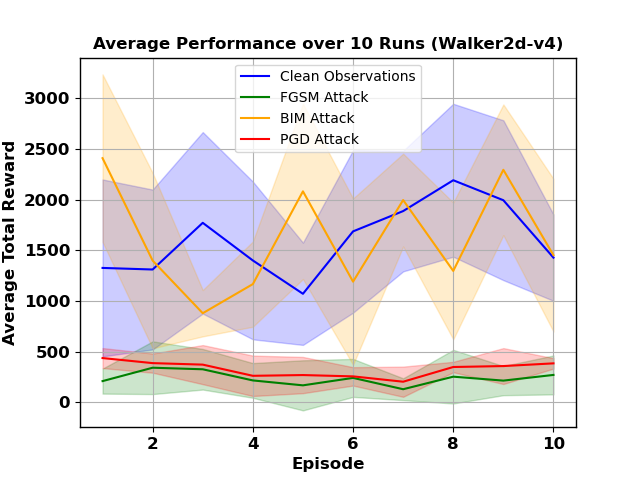}
    \end{minipage}
    \hfill
    \begin{minipage}{0.32\textwidth}
        \includegraphics[width=\textwidth]{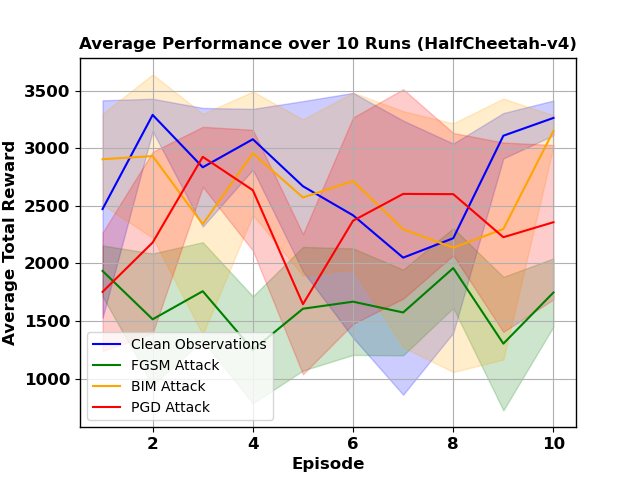}
    \end{minipage}
    \hfill
    \begin{minipage}{0.32\textwidth}
        \includegraphics[width=\textwidth]{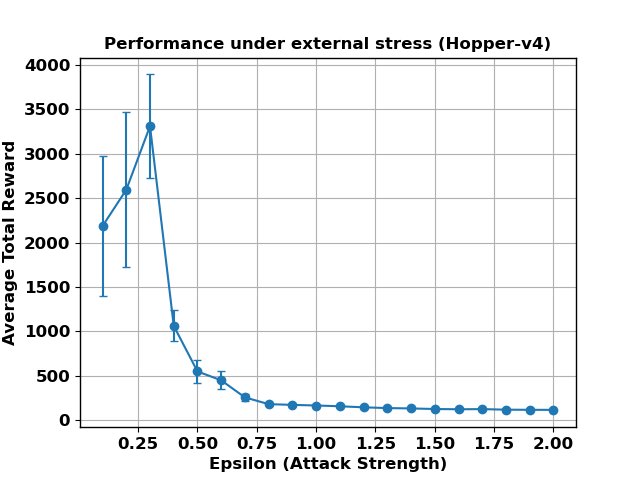}
    \end{minipage}
    \hfill
    \begin{minipage}{0.32\textwidth}
\includegraphics[width=\textwidth]{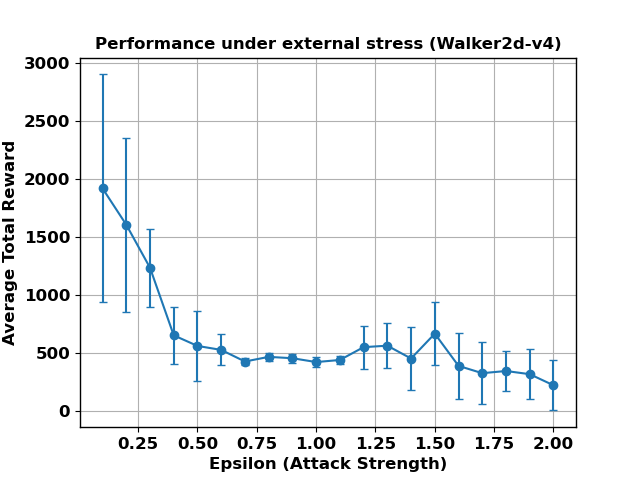}
    \end{minipage}
    \hfill
    \begin{minipage}{0.32\textwidth}
        \includegraphics[width=\textwidth]{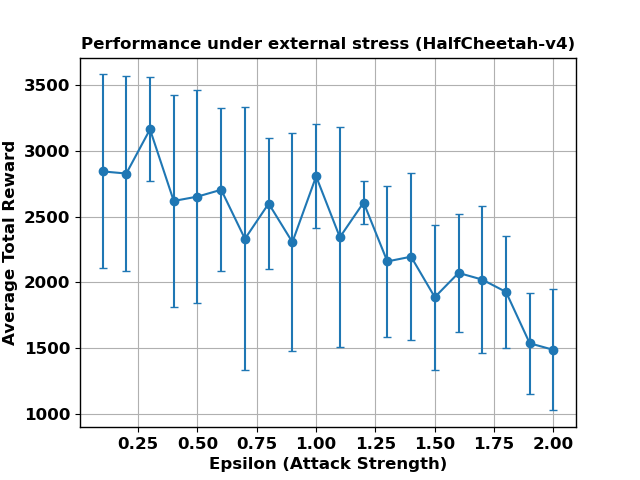}
    \end{minipage}
    \caption{Impact of FGSM, BIM, and PGD Attacks on Policy Performance on Mujoco Environments.}
    \label{adversarialattacks}
\end{figure}
\subsection{Results and Discussions}
Under clean conditions, trained policies achieved stable cumulative rewards across all environments, (see Figure~\ref{adversarialattacks}, top row). The \texttt{Walker2D} policies exhibited the highest baseline rewards of approximately 2000, indicating effective learning and stability. The \texttt{Hopper} policies also performed well but showed slightly higher variability in rewards, suggesting increased sensitivity to parameter perturbations. \texttt{HalfCheetah} policies achieved moderate rewards 2500 with some variability, likely due to its high-dimensional state-action space. These baseline results serve as a reference for assessing how policy performance is affected by internal and external stressors in the subsequent experiments.


\subsubsection{Performance under Adversarial Stress} To evaluate policy robustness, \textcolor{black}{we subjected the trained policies to three gradient-based adversarial attacks includes FGSM \citep{goodfellow2015explaining}, basic iterative method (BIM) \citep{kurakin2018adversarial}, an adversarial attack technique that builds on FGSM by applying it iteratively with small step sizes, and projected gradient descent (PGD)} \citep{madry2018towards}, each applied with varying perturbation magnitudes $\epsilon$. \textcolor{black}{As illustrated in Figure~\ref{adversarialattacks}, FGSM consistently inflicted the most immediate degradation by saturating the $\epsilon$-ball in a single gradient step and causing large reward drops in both \texttt{Walker2D} and \texttt{Hopper}, where returns fell near zero for $\epsilon\geq 0.5$, \textcolor{black}{thereby exposing a critical \emph{vulnerability in the parameter space} and highlighting the \emph{fragility} of these policies to rapid, gradient-driven perturbations.} By contrast, the iterative attacks (BIM and PGD) introduced smaller cumulative shifts due partly to re-sampling and repeated clipping and thus were \emph{less impactful} overall, though they still revealed \emph{robustness limitations} at higher 
$\epsilon$ values. } 
\textcolor{black}{In \texttt{Walker2D}, the policy showed a steady decline in returns with increasing $\epsilon$, stabilizing at~500 for $\epsilon\geq 1.0$.  \texttt{Hopper} demonstrated even greater \emph{sensitivity}, with rewards dropping to near zero at moderate attack strengths  $(\epsilon\geq 0.5)$, highlighting a critical vulnerability in the parameter space. In contrast, \texttt{HalfCheetah} consistently exhibited more \emph{resilience}, retaining moderate rewards around 1500 even under large adversarial shifts $\epsilon= 2.0$, suggesting the presence of \emph{robust} or potentially \emph{antifragile} components in its policy network that adapt to stress and preserve functionality.}
\subsubsection{Performance under internal stress} 
To evaluate the impact of internal stress, synaptic filtering methods--HPF, LPF and PWF--were applied to selectively modify policy parameters. Parameter scores in clean $(S_{\alpha_{i}})$ and adversarial $(S^{\epsilon_{k}})$ conditions were computed to quantify performance changes at various thresholds $\alpha_{i}$. 


\begin{figure}[t]
    \begin{minipage}{0.32\textwidth}
        \includegraphics[width=\textwidth]{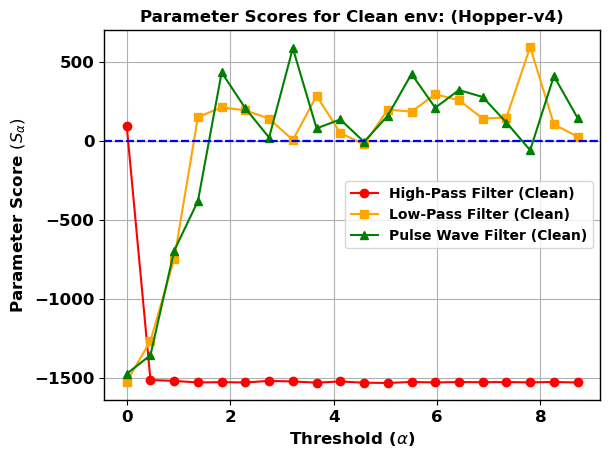}
    \end{minipage}
    \hfill
    \begin{minipage}{0.32\textwidth}
        \includegraphics[width=\textwidth]{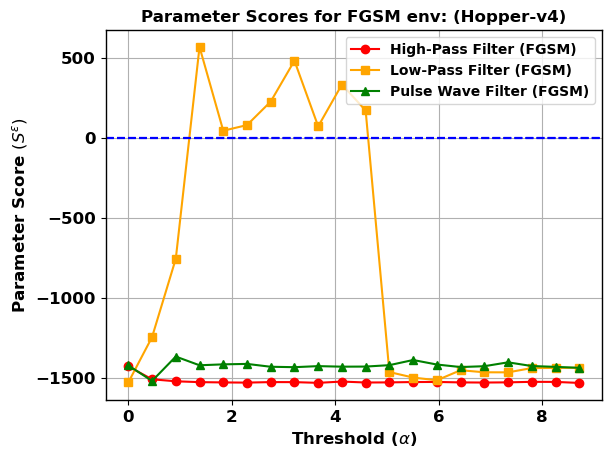}
    \end{minipage}
    \hfill
    \begin{minipage}{0.32\textwidth}
        \includegraphics[width=\textwidth]{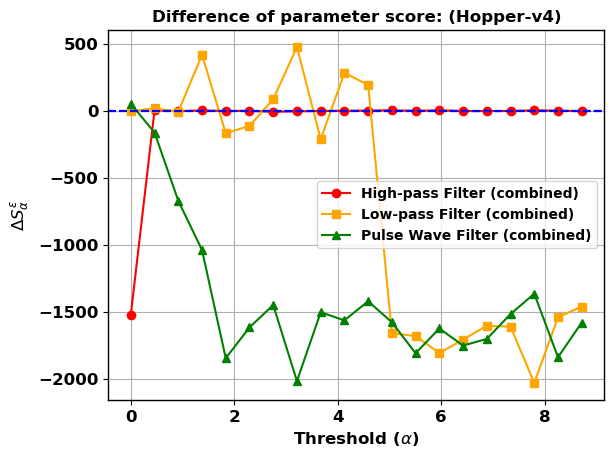}
    \end{minipage}
    \hfill
    \begin{minipage}{0.32\textwidth}
        \includegraphics[width=\textwidth]{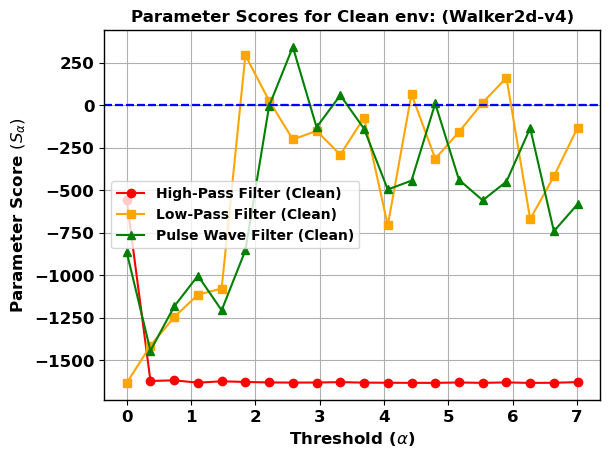}
    \end{minipage}
    \hfill
    \begin{minipage}{0.32\textwidth}
        \includegraphics[width=\textwidth]{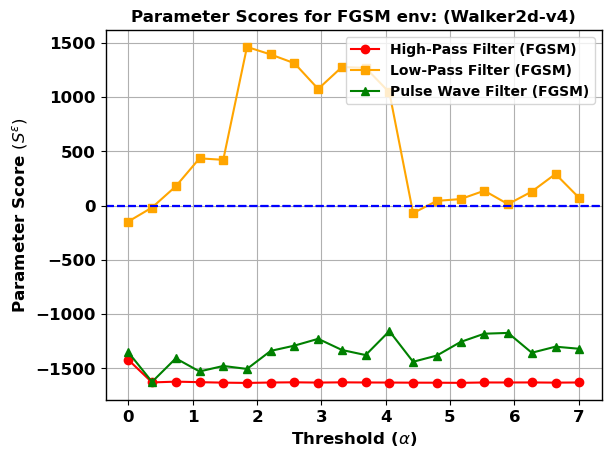}
    \end{minipage}
    \hfill
    \begin{minipage}{0.32\textwidth}
        \includegraphics[width=\textwidth]{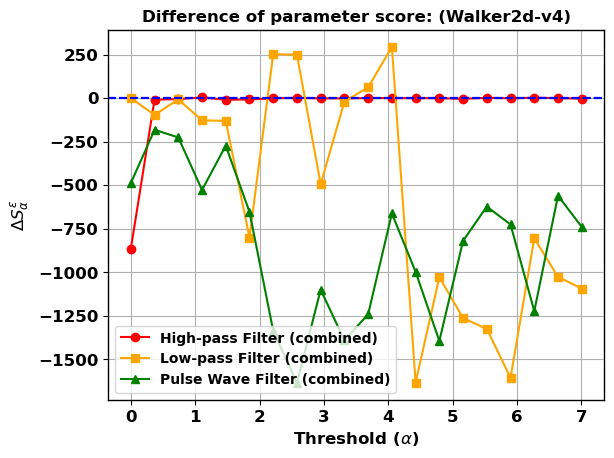}
    \end{minipage}
    \hfill
    \begin{minipage}{0.32\textwidth}
        \includegraphics[width=\textwidth]{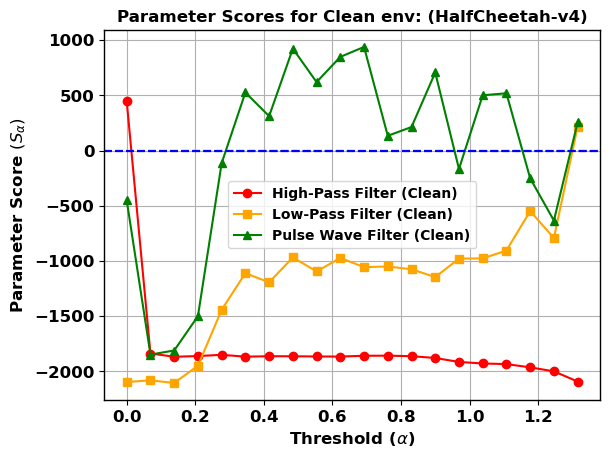}
    \end{minipage}
    \hfill
    \begin{minipage}{0.32\textwidth}
        \includegraphics[width=\textwidth]{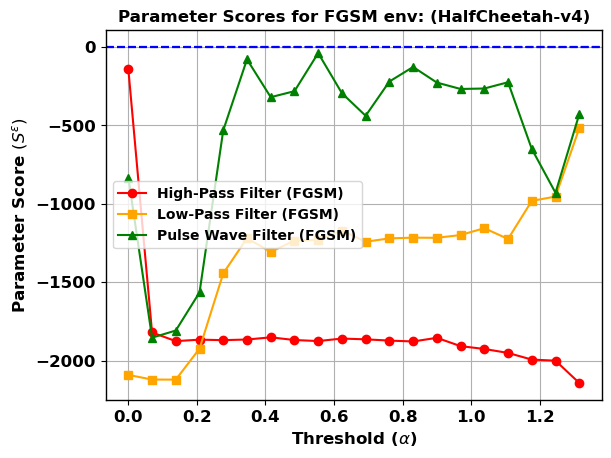}
    \end{minipage}
    \hfill
    \begin{minipage}{0.32\textwidth}
        \includegraphics[width=\textwidth]{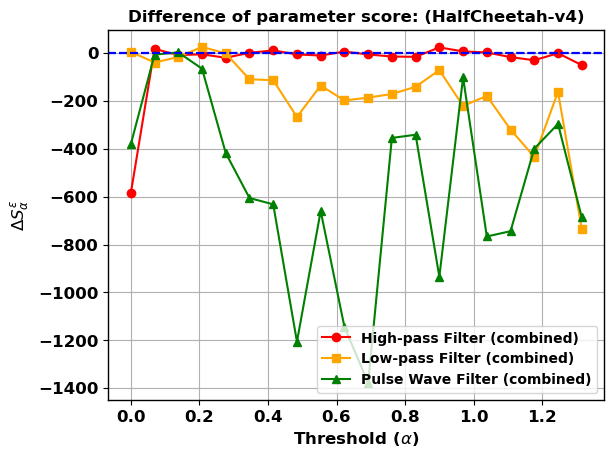}
    \end{minipage}
    \hfill
    \caption{The left column shows parameter scores ($S_{\alpha_i}$) for clean environments. The middle column shows parameter scores ($S^{\epsilon_k}$) under FGSM adversarial perturbations with $\epsilon = 2.0$, while the right column depicts the difference in parameter scores ($\Delta S^{\epsilon_k}_{\alpha_i}$).}
    \label{parameter_score}
\end{figure}

\begin{figure}[t]
    \begin{minipage}{0.32\textwidth}
        \includegraphics[width=\textwidth]{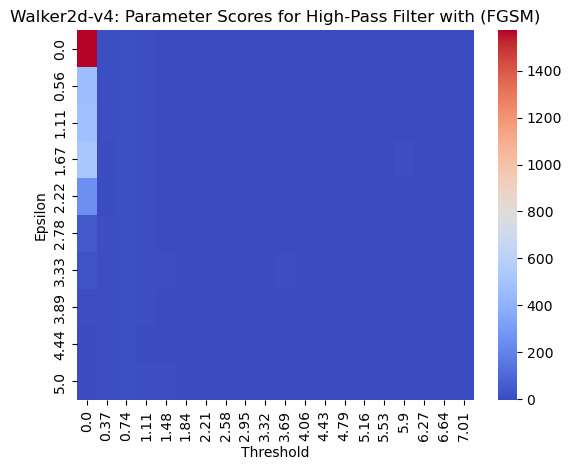}
    \end{minipage}
    \hfill
    \begin{minipage}{0.32\textwidth}
        \includegraphics[width=\textwidth]{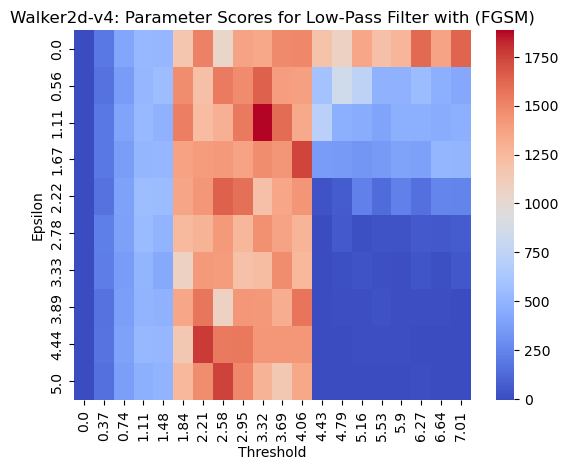}
    \end{minipage}
    \hfill
    \begin{minipage}{0.32\textwidth}
        \includegraphics[width=\textwidth]{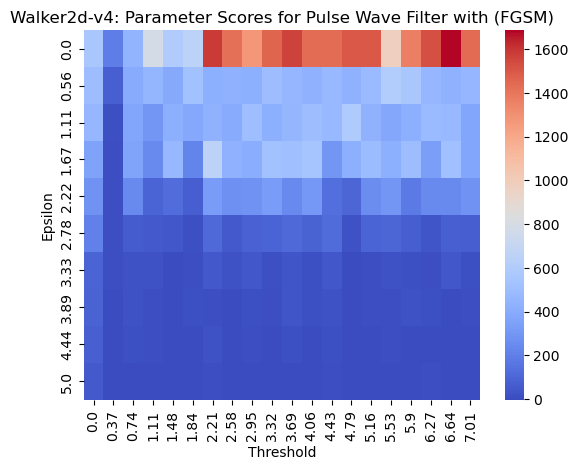}
    \end{minipage}
    \hfill

    \hfill
    \begin{minipage}{0.32\textwidth}
        \includegraphics[width=\textwidth]{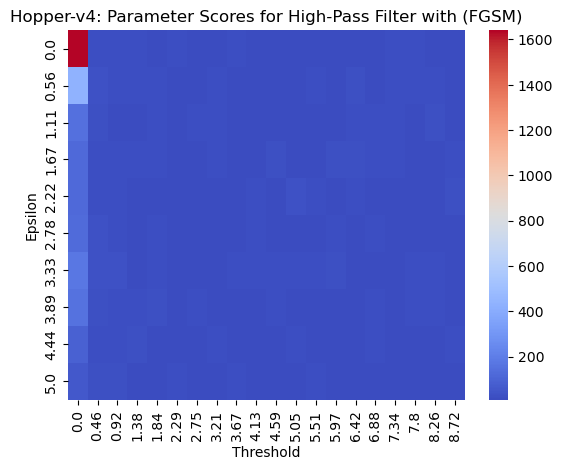}
    \end{minipage}
    \hfill
    \begin{minipage}{0.32\textwidth}
        \includegraphics[width=\textwidth]{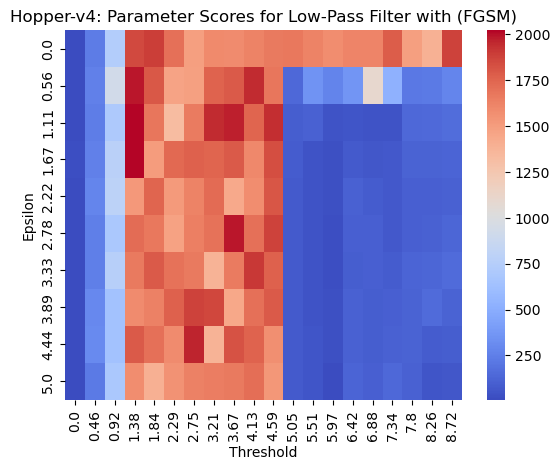}
    \end{minipage}
    \hfill
    \begin{minipage}{0.32\textwidth}
        \includegraphics[width=\textwidth]{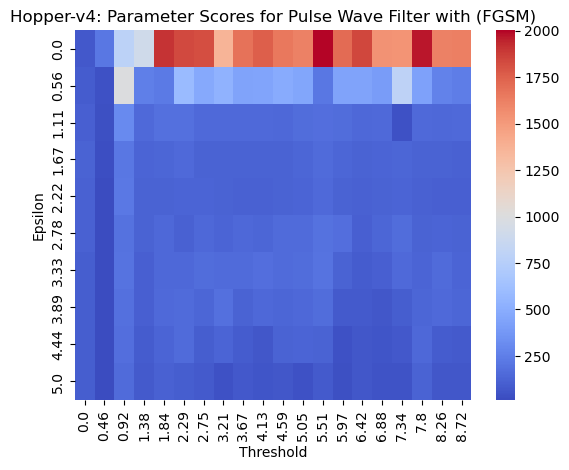}
    \end{minipage}
    \hfill
    \begin{minipage}{0.32\textwidth}
        \includegraphics[width=\textwidth]{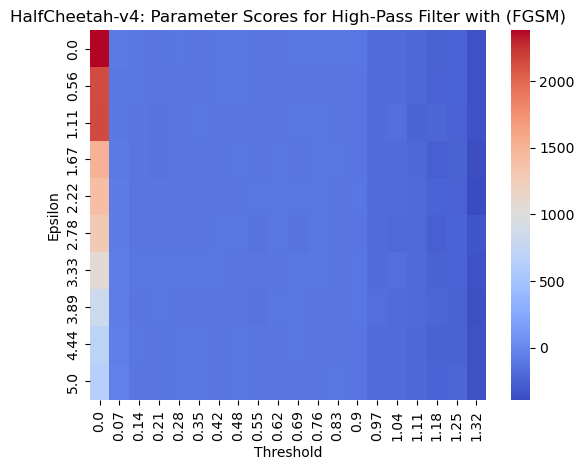}
    \end{minipage}
    \hfill
    \begin{minipage}{0.32\textwidth}
        \includegraphics[width=\textwidth]{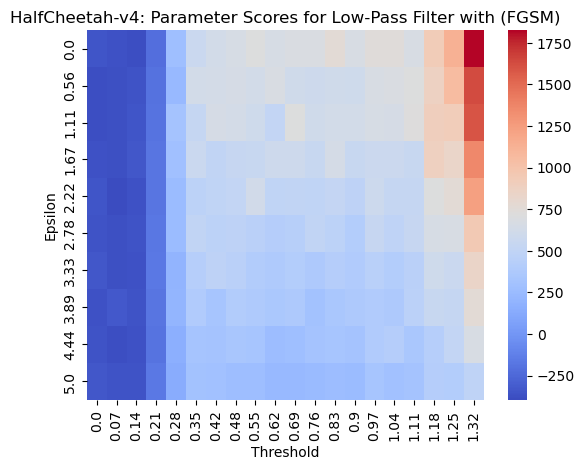}
    \end{minipage}
    \hfill
    \begin{minipage}{0.32\textwidth}
        \includegraphics[width=\textwidth]{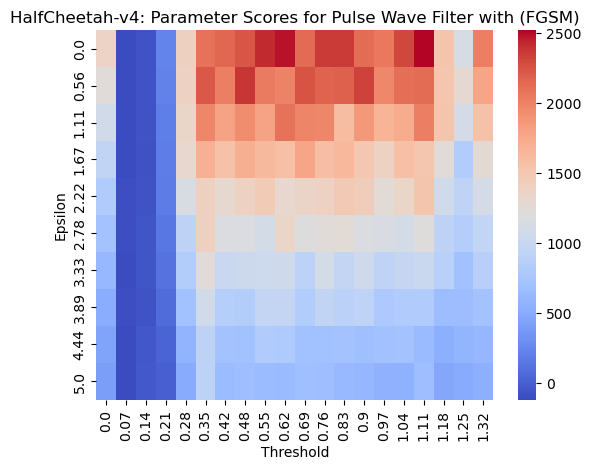}
    \end{minipage}    
    \caption{ Heatmaps of parameter scores under FGSM adversarial attack across environments, showing synaptic filtering methods (High-Pass, Low-Pass, Pulse Wave). The x-axis represents filtering thresholds $\alpha$, and the y-axis denotes stress magnitudes $\epsilon$. Red indicates antifragility, blue indicates fragility.}
    \label{heatmap}
\end{figure}

\textbf{Parameter Scores for Clean Environments:} The left column of Figure~\ref{parameter_score} shows parameter scores $(S_{\alpha_{i}})$ under clean environment. 
Across all environments, high-pass filter consistently yields large negative scores at all thresholds $\alpha$, indicating the presence of \textbf{fragile} parameters, suggests that removing low-magnitude parameters leads to significant degradation in policy performance. Conversely, the low-pass filter shows positive parameter scores at certain thresholds (e.g., $\alpha = 2-4$ in \texttt{Hopper} and \texttt{Walker2D}), indicate \textbf{antifragile} behavior, where removing high-magnitude parameters improves performance. This suggests that dominant parameters are not necessarily beneficial, and pruning them can improve policy execution. The pulse-wave filter exhibits mixed behavior. In \texttt{HalfCheetah}, low thresholds $(0.3\leq\alpha \leq 1.1)$ yields positive scores, highlighting antifragility, while higher thresholds result in negative scores, revealing fragility. This indicates that the pulse wave filter's impact is dependent on the filtering threshold. These results emphasize the utility of low-pass filtering in identifying antifragile parameters.

\textbf{Parameter Scores for Adversarial Environments:} The middle column of Figure \ref{parameter_score}, examine the parameters scores $S^{\epsilon_{k}}$ under adversarial perturbations $\epsilon=2.0$, providing insights into how synaptic filtering interacts with external stress. The high-pass filter continues to show highly negative scores indicating that fragile parameters identified in clean environments remain fragile under adversarial stress. The degradation is further exacerbated by external perturbations. The low-pass filter maintains its antifragility, although the magnitude of positive scores decreases slightly compared to the clean environment. This indicates that parameters identified as \textbf{antifragile} under clean conditions are also \textbf{robust} to adversarial attacks, as evident in \texttt{Hopper} and \texttt{Walker2D}. Increased fragility is observed under adversarial conditions, particularly in \texttt{HalfCheetah}, where most thresholds yield negative scores. This underscores the limited robustness of pulse wave filtering approach. The trends highlight the resilience of low-pass filtering under adversarial stress, maintaining antifragility in key parameter regions.

The heatmaps in Figure~\ref{heatmap} illustrate the variation of parameter scores \( S^{\epsilon_k} \) with internal stress~$(\alpha)$ and external stress (\( \epsilon \)), providing insights into parameter behavior under combined stress conditions. The high-pass filter predominantly identifies fragile parameters, as indicated by consistent performance degradation across the $\alpha$–$\epsilon$ grid. The low-pass filter, in contrast, reveals regions of antifragility, particularly in \texttt{Walker2D}-v4 and \texttt{Hopper}-v4, where moderate thresholds (\( \alpha = 1.84 - 4.06 \), \( \alpha = 0.92 - 4.59 \)) and stress levels (\( \epsilon = 0 - 5 \)) exhibit improved performance due to the removal of high magnitude parameters. These findings suggest that low-pass filtering isolates parameters critical for stability and balance in these environments. In \texttt{HalfCheetah}-v4, antifragility is observed at low thresholds ($\alpha = 1.11$–$1.32$) and moderate adversarial strengths ($\epsilon \leq 2.38$), but performance declines as perturbation increases, indicating lower resilience in this setting. The pulse-wave filter displays heterogeneous patterns across environments. While it exhibits antifragility at mild stress and lower thresholds ($\epsilon \leq 0.2$), it becomes fragile under higher stress levels, suggesting that its effectiveness is highly sensitive to both $\alpha$ and $\epsilon$. Overall, these results highlight the effectiveness of low-pass filtering in identifying parameters that support both robustness and antifragility across diverse environments, particularly where stability and adaptability are essential.

\textbf{Difference of Parameter Scores:} The third column in Figure~\ref{parameter_score} illustrates the difference in parameter scores (\( \Delta S^{\epsilon_k}_{\alpha_i} \)), quantifying the impact of adversarial attacks on synaptic filtering performance. Low-pass filtering shows minimal deviation in \( \Delta S^{\epsilon_k}_{\alpha_i} \), indicating robust parameter identification that performs consistently in both clean and adversarial conditions. In contrast, the pulse-wave filter exhibits high variability, with sharp positive and negative deviations, highlighting its sensitivity to adversarial stress and reduced robustness compared to low-pass filtering. These findings confirm that low-pass filtering is the most effective strategy for isolating stable parameters under combined internal and external stress.

\section{Conclusion and Future Directions}

This study provides a detailed analysis of synaptic filtering on policy network under clean and adversarial conditions. Low-pass filtering consistently identifies robust and antifragile parameters, demonstrating its effectiveness in enhancing network resilience. High-pass filtering highlights fragile parameters that degrade performance, particularly under adversarial stress, while pulse-wave exhibits inconsistent behavior, identifying antifragility only at specific thresholds and showing reduced reliability overall. These findings underscore the importance of targeted filtering strategies to optimize policy performance in stress-prone environments. As a next step, we aim to integrate synaptic filtering directly into the training process, enabling the emergence of parameter structures that not only preserve performance under nominal conditions but also enhance adaptability and resilience against adversarial perturbations.

\end{document}